%% file: sample.tex
\documentclass[twoside,11pt]{article}

\usepackage{blindtext}

\usepackage[preprint]{jmlr2e}
\usepackage{xcolor}         

\usepackage{lastpage}
\jmlrheading{23}{2022}{1-\pageref{LastPage}}{TODO; Revised TODO}{TODO}{21-0000}{Rui Pan, Shizhe Diao, Jianlin Chen and Tong Zhang}

\usepackage{amsmath}
\usepackage{bm}
\usepackage{booktabs}
\usepackage{caption}
\usepackage{makecell}
\usepackage{multirow}
\usepackage{wrapfig}
\usepackage{listings}
\input{ipython.tex}

\newcommand{\papertitle}{\raisebox{0mm}{\includegraphics[width=1.3cm]{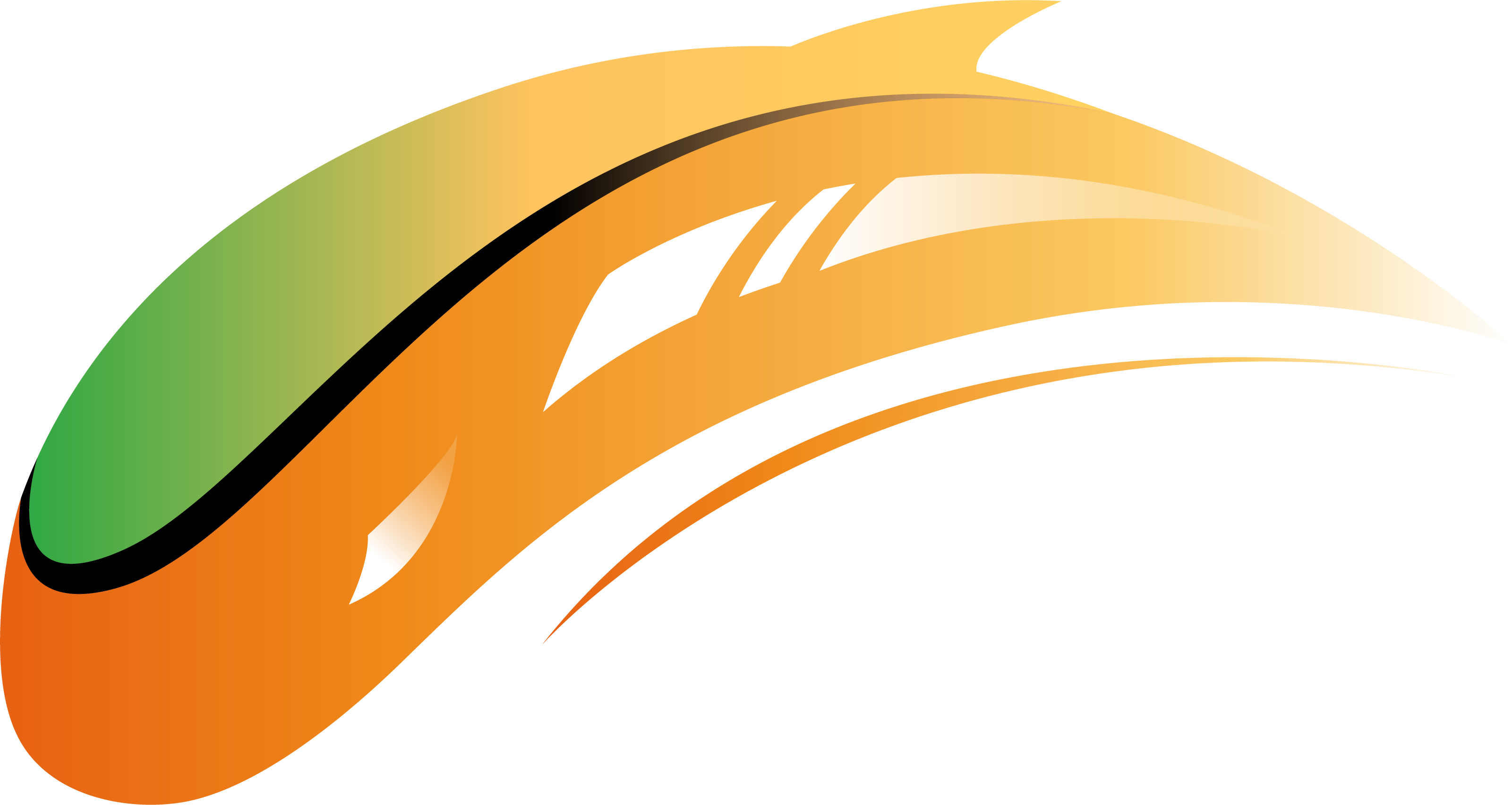}}ExtremeBERT: A Toolkit for Accelerating Pretraining of Customized BERT}

\ShortHeadings{\papertitle}{Pan, Diao, Chen and Zhang}
\firstpageno{1}

\begin{document}

\title{\papertitle}

\author{\name Rui Pan\thanks{Equal contribution.} \email rpan@connect.ust.hk \\
       \addr Department of Computer Science and Engineering \\
       The Hong Kong University of Science and Technology, Hong Kong
\AND
\name Shizhe Diao\footnotemark[1] \email sdiaoaa@connect.ust.hk \\
       \addr Department of Computer Science and Engineering \\
       The Hong Kong University of Science and Technology, Hong Kong
\AND
\name Jianlin Chen\footnotemark[1] \email jc6g20@soton.ac.uk \\
       \addr Department of Social Sciences \\
       University of Southampton, UK
\AND
\name Tong Zhang \email tongzhang@tongzhang-ml.org \\
       \addr Department of Computer Science and Engineering \\
       The Hong Kong University of Science and Technology, Hong Kong
}

\editor{TODO}

\maketitle

\begin{abstract}
In this paper, we present ExtremeBERT, a toolkit for accelerating and customizing BERT pretraining. 
Our goal is to provide an easy-to-use BERT pretraining toolkit for the research community and industry.
Thus, the pretraining of popular language models on customized datasets is affordable with limited resources. Experiments show that, to achieve the same or better GLUE scores, the time cost of our toolkit is over $6\times$ times less for BERT Base and $9\times$ times less for BERT Large when compared with the original BERT paper. 
The documentation and code are released at \url{https://github.com/extreme-bert/extreme-bert} under the Apache-2.0 license.
\end{abstract}

\begin{keywords}
  BERT, Pretraining, Easy-to-Use, Acceleration, Customized Datasets, Language Models.
\end{keywords}

\section{Introduction}

The field of Natural Language Processing (NLP) has made significant progress since the introduction of large language models, such as BERT~\citep{devlin2018bert}, GPT-3~\citep{brown2020gpt3} and T5~\citep{raffel2020T5}.
However, most large language models require a stage of pretraining before being applied to the target NLP task, where the pretraining cost could be staggering.

Several libraries have been proposed to alleviate the above problem. 
For instance, HuggingFace~\citep{wolf-etal-2020-huggingface} \footnote{Code repository: \url{https://github.com/huggingface/transformers}} provides a huge number of pretrained models, allowing users to download the model directly from the model repository. 
It also encourages users to upload their own pretrained models, so that others can utilize the same model without repeating any expensive pretraining process. 
Nevertheless, those features are still insufficient for researchers or engineers with privacy concerns or customization needs. 
For example, in most hospitals or banks, the data has its own value. 
Releasing the model trained on those data will not only compromise user privacy, but can also incur serious security issues. 
Due to this very reason, those pretrained models are no longer available online, forcing one to pretrain the whole model from scratch. 
However, HuggingFace provides slight acceleration for pretraining models under the limited-resource scenario.
Hence the time cost still remains a huge obstacle. 

On the other hand, several other libraries provide significant speed-up for pretraining BERT~\citep{devlin2018bert}. For instance, Academic BERT~\citep{24hbert2021} summarizes a list of tricks to reduce the pretraining time cost, making it possible to surpass BERT Base on GLUE tasks~\citep{wang2018glue} at significantly less cost. 
However, this toolkit restricts itself to the original BERT settings and lacks sufficient flexibility for customized models or datasets. 
Furthermore, its official repository \footnote{Code repository: \url{https://github.com/IntelLabs/academic-budget-bert}} misses an easy-to-use pipeline and requires users to manually build the working environment, modify its codes and run the programs stage by stage in order to reproduce the reported results.

To resolve those pain points, we propose ExtremeBERT, a toolkit that obtains an-order-of-magnitude acceleration in BERT pretraining. 
In the meantime, it supports an easy-to-use pipeline and dataset customization.  

\section{Toolkit Overview}

ExtremeBERT is a toolkit designed to support fast language model pretraining under a limited-resource scenario. 
It can significantly speed up the pretraining process of BERT through combining several recent accelerating techniques. 
Meanwhile, ExtremeBERT supports a wide range of datasets, including customized datasets provided by users and all Huggingface pretraining datasets online \footnote{Huggingface dataset repository: \url{https://huggingface.co/datasets}}. 
Those features are seamlessly incorporated in the toolkit and can be customized simply by specifying a YAML configuration file, where running the whole pipeline only requires a one-line command. 

The whole toolkit is built upon the code repository of~\citet{24hbert2021}, where most of its acceleration features have been inherited. 
On top of that, ExtremeBERT introduces its own acceleration techniques and further improves its already superior speed.
Several missing features and pain points in~\citet{24hbert2021} are also well handled in ExtremeBERT, leading to a much more user-friendly interface. 

\subsection{Feature: Easy-to-use Pipeline}


\begin{figure}[h!]
  \centering
  \includegraphics[scale=0.48]{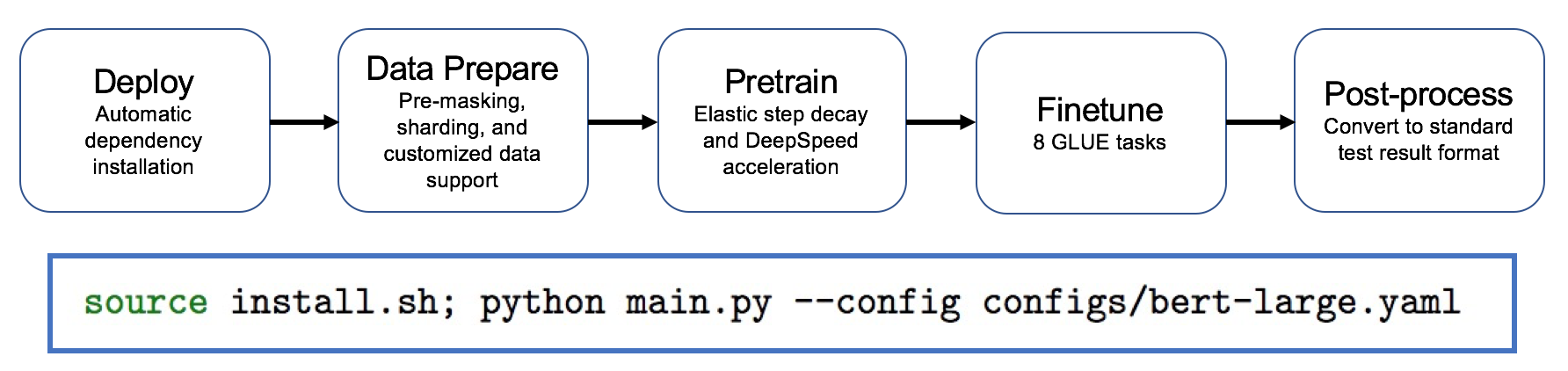}
  \caption{Five stages of the ExtremeBERT pipeline and the one-line command.}
  \label{fig:pipeline}
\end{figure}
\begin{figure}
\centering
\includegraphics[scale=0.6]{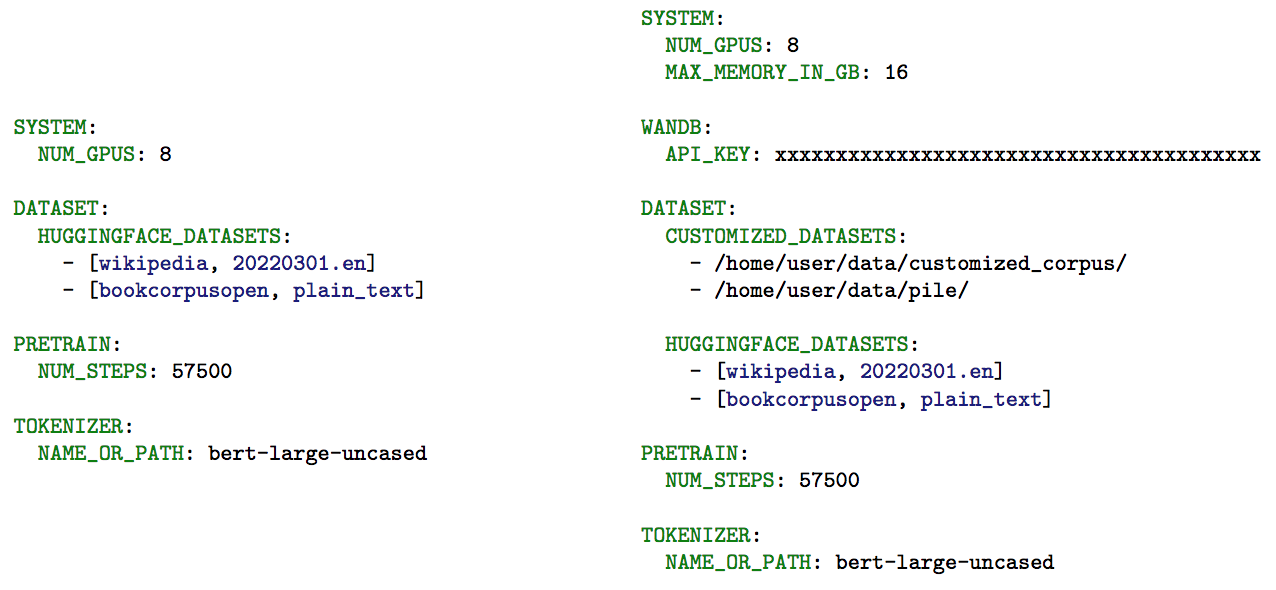}
\caption{\textbf{Left:} An example YAML configuration file for BERT Large \texttt{bert-large.yaml}; \textbf{Right:} An example YAML configuration file for a customized BERT model \texttt{bert-customized.yaml}.}
\label{fig:pipeline_config}
\end{figure}

Many frameworks, including Huggingface~\citep{wolf-etal-2020-huggingface} and Academic BERT~\citep{24hbert2021}, require users to manually type input commands or write codes for each stage of the pipeline. 
However, this is not only painstaking, it is also inefficient. 
This pain point is solved in ExtremeBERT by introducing an one-line command pipeline. 
It asks users to provide a configuration file that only includes the necessary information of BERT pretraining. 

In addition, ExtremeBERT solely relies on python packages, freeing users from complex environment preparation that is common in many other accelerated frameworks, e.g. the APEX C++ toolkit dependency problem in~\citet{24hbert2021}.

\subsection{Feature: Acceleration}


\begin{wrapfigure}{r}{0.3\textwidth}
  \centering
  \includegraphics[scale=0.22]{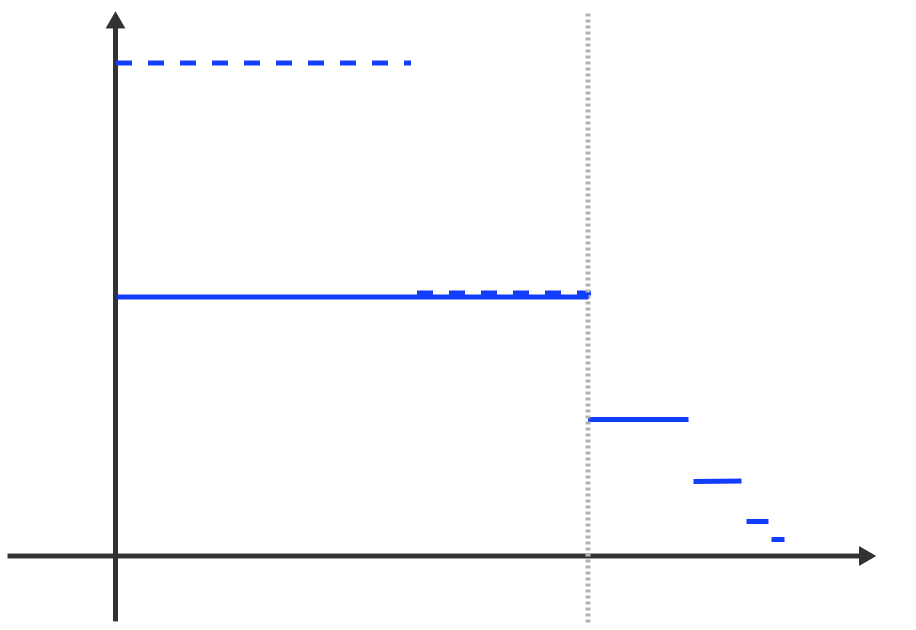}
  \caption{Illustration of the modified ESD scheduler.} 
  \label{fig:esd_in_practice_illust}
\end{wrapfigure}

ExtremeBERT inherits all acceleration techniques of Academic BERT~\citep{24hbert2021}, including the DeepSpeed Package~\citep{rasley2020deepspeed}, mixed-precision training, large batch training, insufficient large model training, STILT tricks~\citep{phang2018stilt} and so on.

On top of that, ExtremeBERT adopts a modified Elastic Step Decay (ESD) learning rate scheduler~\citep{pan2021eigencurve} to further accelerate the pretraining process.

\begin{equation}
\label{eq:esd_in_practice}
\begin{aligned}
  \eta_t
  =&
  \begin{cases}
    \eta_0,
    \quad& t \in \left[0, 1 - r^{\ell}\right] \cdot T,
    \\
    \eta_0 \cdot \left(\frac{1}{2r}\right)^{i - \ell}
    \quad& t \in \left(1-r^{i-1}, 1 - r^{i}\right] \cdot T, i \ge \ell + 1.
  \end{cases}
\end{aligned}
\end{equation}

Experiments show that this technique results in $\sim 0.4$ improvements over the linear decay scheduler in the final GLUE score.

\subsection{Feature: Datasets}

ExtremeBERT supports two type of datasets: customized datasets provided by users and datasets supported by Huggingface. 

For customized datasets, users only need to convert their corpus into raw text files, where in each text file, articles are assumed to be separated by blank lines. Then by grouping those text files in folders and specifying the folders' paths in the pipeline configuration file, ExtremeBERT will automatically handle the data pre-processing on given datasets, i.e. pre-masking, shuffling, sharding and tokenization.

For Huggingface datasets, as shown in Figure~\ref{fig:pipeline_config}, only require the dataset name and split name to be specified in the pipeline configuration file, e.g. \texttt{bookcorpus} and \texttt{plain\_text}, respectively. ExtremeBERT will then download the corresponding datasets from the Huggingface dataset repository and handle the pre-processing.

On top of that, ExtremeBERT supports large dataset processing on small RAM servers. In the toolkit provided by~\citet{24hbert2021}, it is observed that the pre-processing of Wikipedia and bookcorpus datasets costs more than 60GB memory, rendering it impossible for researchers to use them on small servers such as Google Colab, which typically supports $<25$ GB memory. To solve this issue, ExtremeBERT introduces a time-memory tradeoff mechanism to enable small-RAM pre-processing, which basically utilizes disk serialization to save RAM usage. 
By specifying the \texttt{SYSTEM:MAX\_MEMORY\_IN\_GB} field in the pipeline configuration file, as shown in Figure~\ref{fig:pipeline_config}, ExtremeBERT will control the RAM usage for data pre-processing and make sure no out-of-memory issue occurs.

\section{Performance}
\label{sec:exp}

ExtremeBERT provides significant speedup over BERT's original toolkit~\citep{devlin2018bert}. 
To quantify this acceleration, we compare the cost of reaching or surpassing the benchmark GLUE score, as shown in Table~\ref{tab:extremebert_glue_overview}. Here all empirical experiments are conducted on BERT Large models with different hyperparameters, the same as the convention in~\citet{24hbert2021}.
For more experimental details, please refer to Appendix~\ref{appendix:glue_experiments}.

\begin{table}[h!]
  \scriptsize
  \begin{center}
    \begin{tabular}{ccccccc}
      \toprule

      \makecell{Benchmark}
      & Toolkit
      & \makecell{GPU \\ resource}
      & \makecell{GLUE \\ score}
      & \makecell{Pretraining \\ Time (days)}
      & \makecell{Cost\\(V100 $\times$ days)}
      & \textbf{Speedup}

      \\
      \midrule

      \multirow{2}{*}{BERT Base}
      & Original Paper
      & TPU chips $\times 16$
      & 79.6
      & $4.0$
      & $> 64$ 
      & -
      \\



      & ExtremeBERT 
      & GeForce RTX 3090 $\times 8$
      & 81.1
      & $1.3$
      & $\sim 10.4$
      & $\bm{6.1 \times}$
      \\
      \midrule

      \multirow{2}{*}{BERT Large}
      & Original Paper
      & TPU chips $\times 64$
      & 82.1
      & $4.0$
      & $>256$ 
      & -
      \\

      & ExtremeBERT 
      & GeForce RTX 3090 $\times 8$
      & 82.4
      & 3.4
      & $\sim 27.2$
      & $\bm{9.4 \times}$
      \\
      
      \bottomrule
    \end{tabular}
  \end{center}
  \caption{GLUE: results on 9 tasks selected by~\citep{devlin2018bert}.}
  \label{tab:extremebert_glue_overview}
\end{table}

Experiment results demonstrate that our toolkit offers roughly one order of magnitude speedup over the original BERT toolkit. This renders it possible for researchers and engineers to pretrain a BERT model from scratch with limited resources. 

\section{Conclusion and Future Work}


In this paper, we introduce ExtremeBERT, a resource-friendly framework for BERT pretraining, which supports customized datasets and offers roughly one order of magnitude acceleration over the original BERT pretraining process. To obtain the desired customized pretrained model, users are only a configuration file and one-line command away, where the remaining part will be handled smoothly by ExtremeBERT. 

In the future, ExtremeBERT will support more features, including more diverse models, multi-server dataset pre-processing, and SuperGLUE~\citep{wang2019superglue} for finetuning tasks. Our vision is to make the pretraining of language models possible for every researcher and engineer.






\newpage

\appendix
\section{Experimental Details in Section~\ref{sec:exp}}
\label{appendix:glue_experiments}

For the BERT Base benchmark, we follow the convention of~\citet{24hbert2021}, where a BERT Large model is insufficiently trained to obtain the performance of BERT Base. We 
first use $6\%$ portions of iterations to linearly ``warm-up'' the learning rate from $0$ to the peak learning rate $\eta_0 = 10^{-3}$. Then we invoke a modified 
version of Elastic Step Decay~\citep{pan2021eigencurve} for the rest of the iterations
\begin{equation}
\label{eq:esd_in_practice}
\begin{aligned}
  \eta_t
  =&
  \begin{cases}
    \eta_0,
    \quad& t \in \left[0, 1 - r^{\ell}\right] \cdot T,
    \\
    \eta_0 \cdot \left(\frac{1}{2r}\right)^{i - \ell}
    \quad& t \in \left(1-r^{i-1}, 1 - r^{i}\right] \cdot T, i \ge \ell + 1.
  \end{cases}
\end{aligned}
\end{equation}
The default hyperparameters are set to be $\eta_0 = 2 \times 10^{-3}$, $r = 2^{-1/2}$ and $\ell = 6$. As for the number of iterations, we run 23k iterations instead of a 1-day budget, since it is invariant to the machine settings and more reproducible. Other pretraining/finetuning settings remain the same as~\citep{24hbert2021}.

For the BERT Large benchmark, the only difference is the number of iterations and the peak learning rate. For the number of iterations, we use 57.5k iterations instead of 23k iterations, thus the model is more sufficiently trained. For the peak learning rate, we use a smaller learning rate $\eta_0 = 10^{-3}$ to make the training process more stable.

In all experiments, we limit the sequences to 128 tokens and use single-sentence training without next sentence prediction (NSP) to optimize the pretraining process. We also hold out only 0.5\% of the data and compute the validation loss less frequently to increase the training time.

Moreover, regarding the cost estimation in Table~\ref{tab:extremebert_glue_overview}, we refer to the statistics in \url{https://timdettmers.com/2018/10/17/tpus-vs-gpus-for-transformers-bert/} to make the pretraining costs on different hardwares comparable.

\clearpage
\section{User Manual}
\label{appendix:user_manual}

{
\linespread{1.5} \selectfont
\textbf{Version} 0.1.0 \\
\textbf{Date} 2022-12-01 \\
\textbf{Title} A Tooklit for Accelerating Pretraining of Customized BERT \\
\textbf{Author} Rui Pan (\href{mailto:rpan@connect.ust.hk}{rpan@connect.ust.hk}), Shizhe Diao (\href{mailto:sdiaoaa@connect.ust.hk}{sdiaoaa@connect.ust.hk}), Jianlin Chen (\href{mailto:jc6g20@soton.ac.uk}{jc6g20@soton.ac.uk}), Tong Zhang (\href{mailto:tongzhang@tontzhang-ml.org}{tongzhang@tontzhang-ml.org}) \\
\textbf{Maintainer} Rui Pan (\href{mailto:rpan@connect.ust.hk}{rpan@connect.ust.hk}), Shizhe Diao (\href{mailto:sdiaoaa@connect.ust.hk}{sdiaoaa@connect.ust.hk}) \\ 
\textbf{Depends} Python ($\ge 3.6.0$), CUDA ($\ge 11.2$), CUDA Driver ($\ge 460.67$) \\
\textbf{Description} ExtremeBERT is a toolkit that accelerates the pretraining of customized language models on customized datasets, described in the paper ``ExtremeBERT: A Toolkit for Accelerating Pretraining of Customized BERT''.  \\
\textbf{License} Apache-2.0 license \\
\textbf{NeedsCompilation} no \\
\textbf{Repository} \url{https://github.com/extreme-bert/extreme-bert} \\
\textbf{Date/Publication} 2022-12-01 10:00:00 \\
\textbf{System Requirements}
\begin{itemize}
    \item Linux Ubuntu $\ge$ 18.04
    \item At least 20 GB GPU memory, e.g. GPU 3090 $\times$ 1, or 2080Ti GPU $\times$ 2, GeForce RTX 3070 $\times$ 4, etc.
    \item At least 200GB disk space.
\end{itemize}
}

\subsection{Getting started}
\subsubsection{Installation}
\begin{lstlisting}[language=iPython]
$ source install.sh
\end{lstlisting}
\subsubsection{Configurable Pipeline}
First, one may refer to \path{configs/24h-bert-simple.yaml} and make suitable configs for the pipeline, including datasets, number of gpus available, etc. 
Then, by simply running the following command, the whole pipeline will be executed stage by stage,
\begin{lstlisting}[language=iPython]
$ python main.py --config configs/24h-bert-simple.yaml
\end{lstlisting}
This command will run environment installation, dataset prepration, pretraining, finetuning and test result collection one by one and generate the .zip file for GLUE test server submission under \path{output_test_translated/finetune/*/*.zip}. 

To run each stage separately, one may choose to enable or disable each stage in the configuration file. For example,

\begin{figure}[h!]
\centering
\includegraphics[scale=0.63]{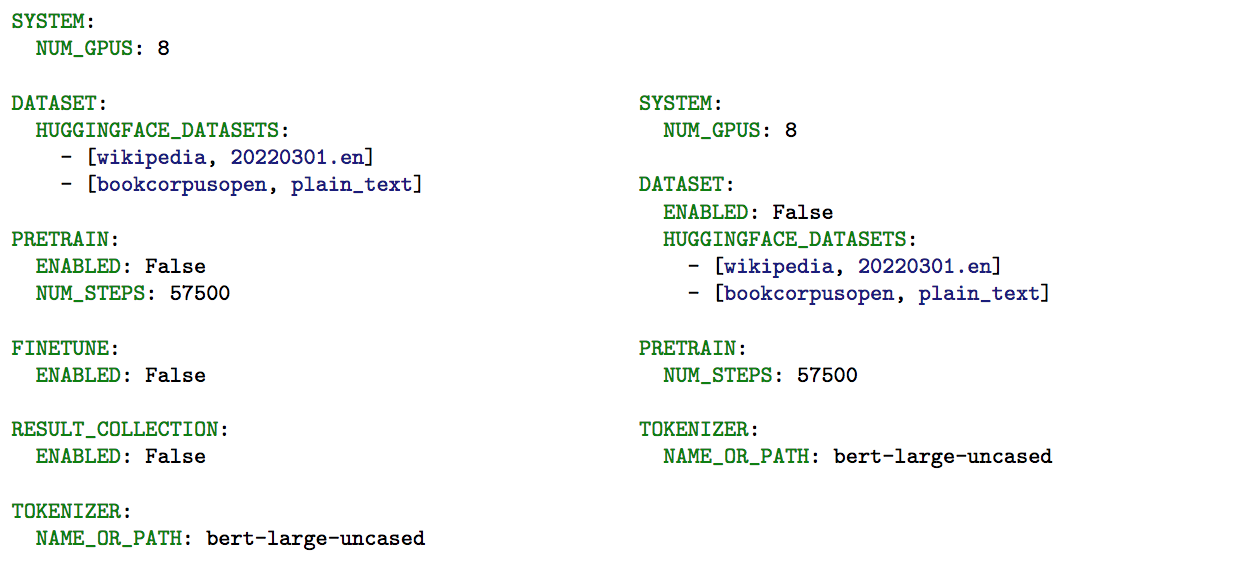}
\caption{\textbf{Left:} An example YAML configuration file \texttt{data-preprocess.yaml} for data preprocessing only ; \textbf{Right:} An example YAML configuration file \texttt{train.yaml} for pretraining + finetuning + result collection when the dataset is preprocessed.}
\label{fig:stage-by-stage_config}
\end{figure}

To obtain more detailed configurations for each stage, one may refer to following sections.

\subsubsection{Preprocessing}
The dataset directory includes scripts to pre-process the datasets we used in our experiments (Wikipedia, Bookcorpus), which can be used as follows. The main pipeline script \path{main.py} is actually invoking those two scripts with different arguments during data preprocessing.
\begin{lstlisting}[language=iPython]
$ python shard_data.py \
  --output_dir {tmp_dir} \
  --num_train_shards 256 \
  --num_test_shards 128 \
  --frac_test 0.1 \
  --max_memory 64 \
  --dataset wikipedia 20220301.en \
  --dataset bookcorpusopen plain_text
\end{lstlisting}

\begin{lstlisting}[language=iPython]
$ python generate_samples.py \
  --dir {tmp_dir} \
  -o {processed_dataset_dir} \
  --dup_factor 10 \
  --seed 42 \
  --do_lower_case 1 \
  --masked_lm_prob 0.15 \ 
  --max_seq_length 128 \
  --tokenizer_name bert-large-uncased \
  --model_name bert-large-uncased \
  --max_predictions_per_seq 20 \
  --n_processes 8
\end{lstlisting}

\subsubsection{Pretraining}
During pretraining, the main pipeline script \path{main.py} first invokes \path{pretrain_search.sh}, which further calls \path{run_pretraining.sh} and finally leads to \path{run_pretraining.py}. For fully configurable settings, please refer to the following command.
\begin{lstlisting}[language=iPython]
$ deepspeed run_pretraining.py \
  --dataset_path {processed_dataset_dir} \
  --output_dir {output_dir} \
  --model_type bert-mlm \
  --tokenizer_name bert-large-uncased \
  --hidden_act gelu \
  --hidden_size 1024 \
  --num_hidden_layers 24 \
  --num_attention_heads 16 \
  --intermediate_size 4096 \
  --hidden_dropout_prob 0.1 \
  --attention_probs_dropout_prob 0.1 \
  --encoder_ln_mode pre-ln \
  --lr 1e-3 \
  --train_batch_size 4096 \
  --train_micro_batch_size_per_gpu 32 \
  --lr_schedule time \
  --curve linear \
  --warmup_proportion 0.06 \
  --gradient_clipping 0.0 \
  --optimizer_type adamw \
  --weight_decay 0.01 \
  --adam_beta1 0.9 \
  --adam_beta2 0.98 \
  --adam_eps 1e-6 \
  --total_training_time 24.0 \
  --early_exit_time_marker 24.0 \
  --print_steps 100 \
  --num_epochs_between_checkpoints 10000 \
  --job_name pretraining_experiment \
  --project_name budget-bert-pretraining \
  --validation_epochs 3 \
  --validation_epochs_begin 1 \
  --validation_epochs_end 1 \
  --validation_begin_proportion 0.05 \
  --validation_end_proportion 0.01 \
  --validation_micro_batch 16 \
  --deepspeed \
  --data_loader_type dist \
  --do_validation \
  --use_early_stopping \
  --early_stop_time 180 \
  --early_stop_eval_loss 6 \
  --seed 42 \
  --fp16
\end{lstlisting}

After pretraining, the pretrained model will be saved to \path{{output_dir}}, which by default is \path{saved_models/pretrain/{dataset_name}} in our pipeline, where \path{{dataset_name}} is automatically generated based on dataset content and will be displayed in the pretraining log.
Users can also manually specify \path{{dataset_name}} in the pipeline configuration file, which corresponds to \path{DATASET.ID}.

In addition, the intermediate logs will have a backup in \path{log/pretrain/{dataset_name}/} for both standard output stream and error stream.

\subsubsection{Finetuning}
During finetuning, the main pipeline script \path{main.py} first invokes \path{finetune_search.sh}, which further calls \path{run_glue.sh} and finally leads to \path{run_glue.py}. For fully configurable settings, please refer to the following command. Please notice that \path{finetune_search.sh} will automatically handle the standard hyperparameter search process and the STILT trick~\citep{phang2018stilt}.

\begin{lstlisting}[language=iPython]
$ python run_glue.py \
  --model_name_or_path {path_to_model_checkpoint} \
  --task_name MRPC \
  --max_seq_length 128 \
  --output_dir /tmp/finetuning \
  --overwrite_output_dir \
  --do_train --do_eval \
  --evaluation_strategy steps \
  --per_device_train_batch_size 32 \
  --gradient_accumulation_steps 1 \
  --per_device_eval_batch_size 32 \
  --learning_rate 5e-5 \
  --weight_decay 0.01 \
  --eval_steps 50 \
  --evaluation_strategy steps \
  --max_grad_norm 1.0 \
  --num_train_epochs 5 \
  --lr_scheduler_type polynomial \
  --warmup_steps 50
\end{lstlisting}
The finetune result of each hyperparameter will be displayed in the log on screen. In our pipeline, this log will be redirected and saved in \path{log/finetune/{dataset_name}/{task_name}/*/}. The test set prediction will also be generated during finetuning, which will be saved in \path{output/finetune/{dataset_name}/{task_name}/*/}. Those files will be parsed later in result collection stage to generate the final validation result and submission zip for \url{https://gluebenchmark.com/}.

\subsection{Functions}

\noindent \rule[0pt]{15.3cm}{0.05em}
BasePretrainModel \textit{Base pretraining model definition}

\noindent \rule[0pt]{15.3cm}{0.05em}

\paragraph{Description}
The base class definition of pretraining model.

\noindent \textbf{Usage}
\begin{lstlisting}[language=iPython]
model = BasePretrainModel(args)
\end{lstlisting}

\paragraph{Arguments}
args: configurations

\paragraph{Value}
an object of pretraining model.

\noindent \rule[0pt]{15.3cm}{0.05em}
get\_default\_config \textit{Get default configuration}

\noindent \rule[0pt]{15.3cm}{0.05em}

\paragraph{Description}
Get default configuration for pipeline training.

\noindent \textbf{Usage}
\begin{lstlisting}[language=iPython]
config = get_default_config()
\end{lstlisting}

\paragraph{Arguments}
N/A

\paragraph{Value}
A dictionary tree is returned, which contains the following components:
\begin{itemize}
    \item config.SYSTEM: system configurations
    \item config.WANDB: wandb configurations
    \item config.DATASET: dataset configurations
    \item config.PRETRAIN: pretrain configurations
    \item config.FINETUNE: finetune configurations
    \item config.TOKENIZER: tokenizer configurations
\end{itemize}

\noindent \rule[0pt]{15.3cm}{0.05em}
prepare\_dataset \textit{Prepare dataset}

\noindent \rule[0pt]{15.3cm}{0.05em}

\paragraph{Description}
Prepare dataset for pretraining.

\noindent \textbf{Usage}
\begin{lstlisting}[language=iPython]
prepare_dataset(config)
\end{lstlisting}

\paragraph{Arguments}
config: configurations

\paragraph{Value}
N/A

\noindent \rule[0pt]{15.3cm}{0.05em}
pretrain \textit{Pretraining entry point}

\noindent \rule[0pt]{15.3cm}{0.05em}

\paragraph{Description}
Pretraining function.

\noindent \textbf{Usage}
\begin{lstlisting}[language=iPython]
pretrain(config)
\end{lstlisting}

\paragraph{Arguments}
config: configurations

\paragraph{Value}
N/A

\noindent \rule[0pt]{15.3cm}{0.05em}
train \textit{Pretraining function}

\noindent \rule[0pt]{15.3cm}{0.05em}

\paragraph{Description}
Pretraining function including iterations over datasets.

\noindent \textbf{Usage}
\begin{lstlisting}[language=iPython]
eval_loss, scale_counter = train(
    args,
    index,
    model,
    optimizer,
    lr_scheduler,
    pretrain_dataset_provider,
    validation_dataset,
)
\end{lstlisting}

\paragraph{Arguments}
\begin{itemize}
    \item args: arguments
    \item index: index of epoch
    \item model: training model
    \item optimizer: optimizer
    \item lr\_scheduler: learning rate scheduler
    \item pretrain\_dataset\_provider: the pretraining dataset
    \item validation\_dataset: validation dataset
\end{itemize}

\paragraph{Value}
\begin{itemize}
    \item eval\_loss: evaluation loss
    \item scale\_counter\_at\_1: if stuck at scale 1, add to scale counter
\end{itemize}

\noindent \rule[0pt]{15.3cm}{0.05em}
finetune \textit{Finetuning entry point}

\noindent \rule[0pt]{15.3cm}{0.05em}

\paragraph{Description}
Finetuning on downstream tasks.

\noindent \textbf{Usage}
\begin{lstlisting}[language=iPython]
finetune(config)
\end{lstlisting}

\paragraph{Arguments}
config: configurations

\paragraph{Value}
N/A

\noindent \rule[0pt]{15.3cm}{0.05em}
collect\_test\_result \textit{Collect test results}

\noindent \rule[0pt]{15.3cm}{0.05em}

\paragraph{Description}
Collect test results.

\noindent \textbf{Usage}
\begin{lstlisting}[language=iPython]
collect_test_result(config)
\end{lstlisting}

\paragraph{Arguments}
config: configurations

\paragraph{Value}
N/A

\paragraph{Details}
It will run `summarize\_val.sh' to summarize the validation results for each downstream tasks, `collect\_best\_val.sh' to collect best validation results, `translate\_test\_result.sh' to translate the prediction to the desired format according to GLUE benchmark website.

\begin{lstlisting}[language=iPython]
dataset_name = config.DATASET.ID
run_bash(f'./summarize_val.sh {dataset_name}')
run_bash(f'./collect_best_val.sh {dataset_name}')
run_bash(f'./translate_test_result.sh {dataset_name}')
\end{lstlisting}

\bibliography{sample}

\end{document}

%% file: ipython.tex
\definecolor{maroon}{cmyk}{0, 0.87, 0.68, 0.32}
\definecolor{halfgray}{gray}{0.55}
\definecolor{ipython_frame}{RGB}{207, 207, 207}
\definecolor{ipython_bg}{RGB}{247, 247, 247}
\definecolor{ipython_red}{RGB}{186, 33, 33}
\definecolor{ipython_green}{RGB}{0, 128, 0}
\definecolor{ipython_cyan}{RGB}{64, 128, 128}
\definecolor{ipython_purple}{RGB}{170, 34, 255}

\usepackage{listings}
\lstdefinelanguage{iPython}{
    morekeywords={access,and,break,class,continue,def,del,elif,else,except,exec,finally,for,from,global,if,import,in,is,lambda,not,or,pass,print,raise,return,try,while},%
    morekeywords=[2]{abs,all,any,basestring,bin,bool,bytearray,callable,chr,classmethod,cmp,compile,complex,delattr,dict,dir,divmod,enumerate,eval,execfile,file,filter,float,format,frozenset,getattr,globals,hasattr,hash,help,hex,id,input,int,isinstance,issubclass,iter,len,list,locals,long,map,max,memoryview,min,next,object,oct,open,ord,pow,property,range,raw_input,reduce,reload,repr,reversed,round,set,setattr,slice,sorted,staticmethod,str,sum,super,tuple,type,unichr,unicode,vars,xrange,zip,apply,buffer,coerce,intern},%
    sensitive=true,%
    morecomment=[l]\#,%
    morestring=[b]',%
    morestring=[b]",%
    morestring=[s]{'''}{'''},
    morestring=[s]{"""}{"""},
    morestring=[s]{r'}{'},
    morestring=[s]{r"}{"},%
    morestring=[s]{r'''}{'''},%
    morestring=[s]{r"""}{"""},%
    morestring=[s]{u'}{'},
    morestring=[s]{u"}{"},%
    morestring=[s]{u'''}{'''},%
    morestring=[s]{u"""}{"""},%
    literate=
    {á}{{\'a}}1 {é}{{\'e}}1 {í}{{\'i}}1 {ó}{{\'o}}1 {ú}{{\'u}}1
    {Á}{{\'A}}1 {É}{{\'E}}1 {Í}{{\'I}}1 {Ó}{{\'O}}1 {Ú}{{\'U}}1
    {à}{{\`a}}1 {è}{{\`e}}1 {ì}{{\`i}}1 {ò}{{\`o}}1 {ù}{{\`u}}1
    {À}{{\`A}}1 {È}{{\'E}}1 {Ì}{{\`I}}1 {Ò}{{\`O}}1 {Ù}{{\`U}}1
    {ä}{{\"a}}1 {ë}{{\"e}}1 {ï}{{\"i}}1 {ö}{{\"o}}1 {ü}{{\"u}}1
    {Ä}{{\"A}}1 {Ë}{{\"E}}1 {Ï}{{\"I}}1 {Ö}{{\"O}}1 {Ü}{{\"U}}1
    {â}{{\^a}}1 {ê}{{\^e}}1 {î}{{\^i}}1 {ô}{{\^o}}1 {û}{{\^u}}1
    {Â}{{\^A}}1 {Ê}{{\^E}}1 {Î}{{\^I}}1 {Ô}{{\^O}}1 {Û}{{\^U}}1
    {œ}{{\oe}}1 {Œ}{{\OE}}1 {æ}{{\ae}}1 {Æ}{{\AE}}1 {ß}{{\ss}}1
    {ç}{{\c c}}1 {Ç}{{\c C}}1 {ø}{{\o}}1 {å}{{\r a}}1 {Å}{{\r A}}1
    {€}{{\EUR}}1 {£}{{\pounds}}1
    {^}{{{\color{ipython_purple}\^{}}}}1
    {=}{{{\color{ipython_purple}=}}}1
    {+}{{{\color{ipython_purple}+}}}1
    {*}{{{\color{ipython_purple}$^\ast$}}}1
    {/}{{{\color{ipython_purple}/}}}1
    {+=}{{{+=}}}1
    {-=}{{{-=}}}1
    {*=}{{{$^\ast$=}}}1
    {/=}{{{/=}}}1,
    literate=
    *{-}{{{\color{ipython_purple}-}}}1
     {?}{{{\color{ipython_purple}?}}}1,
    identifierstyle=\color{black}\ttfamily,
    commentstyle=\color{ipython_cyan}\ttfamily,
    stringstyle=\color{ipython_red}\ttfamily,
    keepspaces=true,
    showspaces=false,
    showstringspaces=false,
    rulecolor=\color{ipython_frame},
    frame=single,
    frameround={t}{t}{t}{t},
    framexleftmargin=6mm,
    numbers=left,
    numberstyle=\tiny\color{halfgray},
    backgroundcolor=\color{ipython_bg},
    basicstyle=\footnotesize\ttfamily,
    keywordstyle=\color{ipython_green}\ttfamily,
    aboveskip=1.2em,
    belowskip=1.2em,
}

%% file: sample.bbl
\begin{thebibliography}{10}
\providecommand{\natexlab}[1]{#1}
\providecommand{\url}[1]{\texttt{#1}}
\expandafter\ifx\csname urlstyle\endcsname\relax
  \providecommand{\doi}[1]{doi: #1}\else
  \providecommand{\doi}{doi: \begingroup \urlstyle{rm}\Url}\fi

\bibitem[Brown et~al.(2020)Brown, Mann, Ryder, Subbiah, Kaplan, Dhariwal,
  Neelakantan, Shyam, Sastry, Askell, Agarwal, Herbert-Voss, Krueger, Henighan,
  Child, Ramesh, Ziegler, Wu, Winter, Hesse, Chen, Sigler, Litwin, Gray, Chess,
  Clark, Berner, McCandlish, Radford, Sutskever, and Amodei]{brown2020gpt3}
Tom Brown, Benjamin Mann, Nick Ryder, Melanie Subbiah, Jared~D Kaplan, Prafulla
  Dhariwal, Arvind Neelakantan, Pranav Shyam, Girish Sastry, Amanda Askell,
  Sandhini Agarwal, Ariel Herbert-Voss, Gretchen Krueger, Tom Henighan, Rewon
  Child, Aditya Ramesh, Daniel Ziegler, Jeffrey Wu, Clemens Winter, Chris
  Hesse, Mark Chen, Eric Sigler, Mateusz Litwin, Scott Gray, Benjamin Chess,
  Jack Clark, Christopher Berner, Sam McCandlish, Alec Radford, Ilya Sutskever,
  and Dario Amodei.
\newblock Language models are few-shot learners.
\newblock In H.~Larochelle, M.~Ranzato, R.~Hadsell, M.F. Balcan, and H.~Lin,
  editors, \emph{Advances in Neural Information Processing Systems}, volume~33,
  pages 1877--1901. Curran Associates, Inc., 2020.
\newblock URL
  \url{https://proceedings.neurips.cc/paper/2020/file/1457c0d6bfcb4967418bfb8ac142f64a-Paper.pdf}.

\bibitem[Devlin et~al.(2018)Devlin, Chang, Lee, and Toutanova]{devlin2018bert}
Jacob Devlin, Ming-Wei Chang, Kenton Lee, and Kristina Toutanova.
\newblock Bert: Pre-training of deep bidirectional transformers for language
  understanding.
\newblock \emph{arXiv preprint arXiv:1810.04805}, 2018.

\bibitem[Izsak et~al.(2021)Izsak, Berchansky, and Levy]{24hbert2021}
Peter Izsak, Moshe Berchansky, and Omer Levy.
\newblock How to train bert with an academic budget, 2021.
\newblock URL \url{https://arxiv.org/abs/2104.07705}.

\bibitem[Pan et~al.(2021)Pan, Ye, and Zhang]{pan2021eigencurve}
Rui Pan, Haishan Ye, and Tong Zhang.
\newblock Eigencurve: Optimal learning rate schedule for {SGD} on quadratic
  objectives with skewed hessian spectrums.
\newblock \emph{CoRR}, abs/2110.14109, 2021.
\newblock URL \url{https://arxiv.org/abs/2110.14109}.

\bibitem[Phang et~al.(2018)Phang, F{\'e}vry, and Bowman]{phang2018stilt}
Jason Phang, Thibault F{\'e}vry, and Samuel~R Bowman.
\newblock Sentence encoders on stilts: Supplementary training on intermediate
  labeled-data tasks.
\newblock \emph{arXiv preprint arXiv:1811.01088}, 2018.

\bibitem[Raffel et~al.(2020)Raffel, Shazeer, Roberts, Lee, Narang, Matena,
  Zhou, Li, and Liu]{raffel2020T5}
Colin Raffel, Noam Shazeer, Adam Roberts, Katherine Lee, Sharan Narang, Michael
  Matena, Yanqi Zhou, Wei Li, and Peter~J. Liu.
\newblock Exploring the limits of transfer learning with a unified text-to-text
  transformer.
\newblock \emph{Journal of Machine Learning Research}, 21\penalty0
  (140):\penalty0 1--67, 2020.
\newblock URL \url{http://jmlr.org/papers/v21/20-074.html}.

\bibitem[Rasley et~al.(2020)Rasley, Rajbhandari, Ruwase, and
  He]{rasley2020deepspeed}
Jeff Rasley, Samyam Rajbhandari, Olatunji Ruwase, and Yuxiong He.
\newblock Deepspeed: System optimizations enable training deep learning models
  with over 100 billion parameters.
\newblock In \emph{Proceedings of the 26th ACM SIGKDD International Conference
  on Knowledge Discovery \& Data Mining}, pages 3505--3506, 2020.

\bibitem[Wang et~al.(2018)Wang, Singh, Michael, Hill, Levy, and
  Bowman]{wang2018glue}
Alex Wang, Amanpreet Singh, Julian Michael, Felix Hill, Omer Levy, and Samuel~R
  Bowman.
\newblock Glue: A multi-task benchmark and analysis platform for natural
  language understanding.
\newblock \emph{arXiv preprint arXiv:1804.07461}, 2018.

\bibitem[Wang et~al.(2019)Wang, Pruksachatkun, Nangia, Singh, Michael, Hill,
  Levy, and Bowman]{wang2019superglue}
Alex Wang, Yada Pruksachatkun, Nikita Nangia, Amanpreet Singh, Julian Michael,
  Felix Hill, Omer Levy, and Samuel Bowman.
\newblock Superglue: A stickier benchmark for general-purpose language
  understanding systems.
\newblock \emph{Advances in neural information processing systems}, 32, 2019.

\bibitem[Wolf et~al.(2020)Wolf, Debut, Sanh, Chaumond, Delangue, Moi, Cistac,
  Rault, Louf, Funtowicz, Davison, Shleifer, von Platen, Ma, Jernite, Plu, Xu,
  Scao, Gugger, Drame, Lhoest, and Rush]{wolf-etal-2020-huggingface}
Thomas Wolf, Lysandre Debut, Victor Sanh, Julien Chaumond, Clement Delangue,
  Anthony Moi, Pierric Cistac, Tim Rault, Rémi Louf, Morgan Funtowicz, Joe
  Davison, Sam Shleifer, Patrick von Platen, Clara Ma, Yacine Jernite, Julien
  Plu, Canwen Xu, Teven~Le Scao, Sylvain Gugger, Mariama Drame, Quentin Lhoest,
  and Alexander~M. Rush.
\newblock Transformers: State-of-the-art natural language processing.
\newblock In \emph{Proceedings of the 2020 Conference on Empirical Methods in
  Natural Language Processing: System Demonstrations}, pages 38--45, Online,
  October 2020. Association for Computational Linguistics.
\newblock URL \url{https://www.aclweb.org/anthology/2020.emnlp-demos.6}.

\end{thebibliography}
